\pdfoutput=1
\documentclass[11pt]{article}

\usepackage[preprint]{acl}

\usepackage{times}
\usepackage{latexsym}

\usepackage[T1]{fontenc}
\usepackage[utf8]{inputenc}

\usepackage{microtype}

\usepackage{inconsolata}

\usepackage{graphicx}

\usepackage{hyperref}       % hyperlinks
\usepackage{url}            % simple URL typesetting
\usepackage{booktabs}       % professional-quality tables
\usepackage{amsfonts}       % blackboard math symbols
\usepackage{nicefrac}       % compact symbols for 1/2, etc.

\usepackage[dvipsnames]{xcolor}
\usepackage{subcaption}
\usepackage{graphicx}
\usepackage{arydshln}
\usepackage{multirow}
\usepackage{wrapfig}
\usepackage{algorithm}
\usepackage[noend]{algpseudocode}
\usepackage{pifont}

\usepackage{enumitem}  % Add this in the preamble if not already included

\usepackage{amsmath}
\usepackage{bm}  % for bold math symbols like \bm{x}
\usepackage[table]{xcolor}  % Put this in the preamble
\usepackage{array}          % For m{} column type

\hypersetup{
    colorlinks,
    linkcolor={red!50!black},
    citecolor={blue!50!black},
    urlcolor={blue!80!black}
}

\usepackage[most]{tcolorbox} % define prompt box
\newtcolorbox[list inside=prompt,auto counter]{prompt}[1][]{
    colbacktitle=black!60,
    coltitle=white,
    fontupper=\footnotesize,
    boxsep=5pt,
    left=0pt,
    right=0pt,
    top=0pt,
    bottom=0pt,
    boxrule=1pt,
    #1,
}

\usepackage{xspace} % set up short-form macros
\newcommand{\model}{\textsc{WebAgent-R1}\xspace}
\newcommand{\modelzero}{\textsc{WebAgent-R1-Zero}\xspace}
\newcommand{\modelcot}{\textsc{WebAgent-R1-CoT}\xspace}

\newcommand{\ie}{{\sl i.e.}}
\newcommand{\eg}{{\sl e.g.}}

\newcommand{\cmark}{\ding{51}}
\newcommand{\xmark}{\ding{56}}

\title{\model: Training Web Agents via End-to-End Multi-Turn Reinforcement Learning}
\author{Zhepei Wei$^{\dag}$\textsuperscript{*}, Wenlin Yao$^\ddag$, Yao Liu$^\ddag$, Weizhi Zhang$^{\ddag}$, Qin Lu$^\ddag$, Liang Qiu$^\ddag$ \\ {\bf Changlong Yu$^\ddag$, Puyang Xu$^\ddag$, Chao Zhang$^\S$, Bing Yin$^\ddag$, Hyokun Yun$^\ddag$, Lihong Li$^\ddag$ }\\ 
$^\dag$University of Virginia \quad
$^\ddag$Amazon \quad
$^\S$Georgia Institute of Technology \\
\texttt{zhepei.wei@virginia.edu, chaozhang@gatech.edu} \\
\texttt{\{ywenlin, yaoliuai, zhweizhi, luqn, liangqxx, changlyu}\\
\texttt{puyax, alexbyin, yunhyoku, llh\}@amazon.com}
}

\begin{document}
\maketitle

\begingroup
\renewcommand\thefootnote{*}
\footnotetext{Work done during internship at Amazon.}
\endgroup

\begin{abstract}
While reinforcement learning (RL) has demonstrated remarkable success in enhancing large language models (LLMs), it has primarily focused on single-turn tasks such as solving math problems.
Training effective web agents for multi-turn interactions remains challenging due to the complexity of long-horizon decision-making across dynamic web interfaces.
In this work, we present \model, a simple yet effective end-to-end multi-turn RL framework for training web agents.
It learns directly from online interactions with web environments by generating diverse trajectories in parallel, entirely guided by binary rewards depending on task success.
Experiments on the WebArena-Lite benchmark demonstrate the effectiveness of \model, boosting the task success rate of Qwen-2.5-3B from 6.1\% to 33.9\% and Llama-3.1-8B from 8.5\% to 44.8\%, significantly outperforming existing state-of-the-art methods and strong proprietary models such as OpenAI o3.
In-depth analyses reveal the effectiveness of the thinking-based prompting strategy and test-time scaling through increased interactions for web tasks.
We further investigate different RL initialization policies by introducing two variants, namely \modelzero and \modelcot, which highlight the importance of the warm-up training stage (\ie, behavior cloning) and provide insights on incorporating long chain-of-thought (CoT) reasoning in web agents.\footnote{Code and artifacts are available at \url{https://github.com/weizhepei/WebAgent-R1}}
\end{abstract}

\section{Introduction}

Reinforcement learning (RL) has emerged as a promising approach for training large language models (LLMs), as exemplified by recent advances such as DeepSeek-R1~\citep{guo2025deepseek, team2025kimi, yang2025qwen3technicalreport}.
However, existing works have primarily focused on single-turn, non-interactive tasks such as mathematical reasoning~\citep{shao2024deepseekmath,zeng2025simplerl}.
Their effectiveness in multi-turn, interactive environments---particularly in complex scenarios requiring long-horizon decision-making and domain-specific skills, such as web browsing~\citep{zhou2024webarena, he2024webvoyager,chae2025web}---still remains underexplored.

Unlike static environments, web tasks pose unique challenges for LLM agents due to their dynamic nature and diverse solution spaces.
Early works on web agents primarily relied on prompting-based methods~\cite{wang2024agent, sodhi2024step, fu2024autoguide, zhang2025webpilot, yang2025agentoccam} or behavior cloning (BC), which imitates demonstrated trajectories via supervised fine-tuning~\citep{yin2024agent, hong2024cogagent, lai2024autowebglm, he2024openwebvoyager, putta2024agent}. Despite their initial success, these methods lack the ability to explore diverse strategies or learn from trial and error, limiting the generalizability of web agents.
To address this issue, recent works explored applying RL for better policy training.
However, most of this line of research has heavily relied on offline or iterative off-policy RL solutions~\citep{peng2019advantage,pan2024autonomous,qi2025webrl}, which break the end-to-end interaction between the web agent and environment, and introduce additional complexities such as trajectory filtering~\citep{bai2024digirl}, outcome reward model training~\citep{qi2025webrl}, or iterative optimization procedures~\citep{zhou2024archer}.
These constraints hinder their practicality for real-world deployment.

Meanwhile, several concurrent works have explored end-to-end RL with on-policy updates for training LLM agents in multi-turn interactive scenarios, such as simulated games and coding environments~\citep{wang2025ragen, cao2025skyrl}.
Unlike off-policy RL that trains on data generated by older versions of the agent, on-policy RL collects training data directly from the agent’s current behavior. This ensures that the learning process is better aligned with the agent’s most recent actions, often leading to more stable and effective learning~\citep{schulman2015high,schulman2017proximal}.
It also eliminates the need for additional overheads in off-policy RL (\eg, maintaining a replay buffer and filtering outdated trajectories), and enables the agent to behave adaptively based on its \emph{own} past decisions---a key advantage in interactive environments where early decisions can significantly affect next steps.

These benefits are particularly desirable in online web environments, which often involve complex interplay between tasks due to dynamic changes of the environment.
For instance, consider a situation where the agent is first tasked to log out of a user account and then to edit the user’s profile.
These tasks are inherently interdependent: once the agent logs out, it loses access to the profile page. If the agent is trained using off-policy data collected from an earlier version that never logged out, it has no opportunity to learn the login behavior and may incorrectly assume continued access and generate invalid actions, ultimately leading to task failure.
End-to-end RL helps avoid such pitfalls by allowing the agent to learn proper behaviors in response to environmental state changes on-the-fly.

\begin{figure}[!t]
\includegraphics[width=0.48\textwidth]{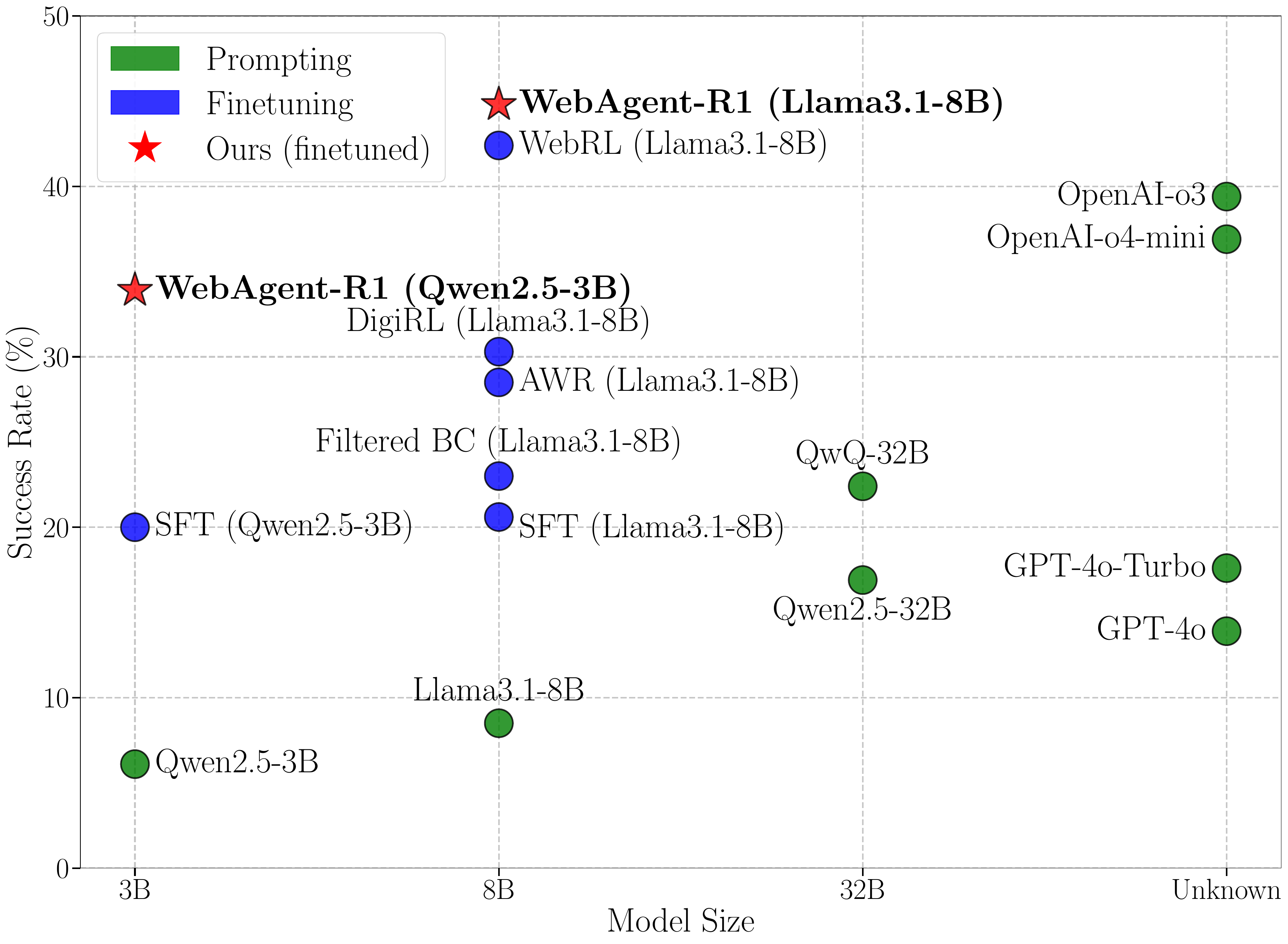}
\vspace{-2em}
\caption{Comparison between existing methods and our \model{} on the WebArena-Lite benchmark. Our method outperforms both strong prompting-based and finetuned baselines, achieving superior performance across various model sizes.}
\label{fig:benchmark_overview}
\vspace{-1.5em}
\end{figure}

In light of this, we propose \model, an end-to-end multi-turn RL framework for training web agents.
Specifically, our design addresses several key challenges in this setting.
First, at each step, the environmental observation (\eg, HTML content) can span thousands of tokens, causing the accumulated context over long horizons to incur substantial memory overheads.
To mitigate this, we introduce a dynamic context compression mechanism, which adaptively adjusts the contexts across turns, ensuring scalability and preventing out-of-memory issues.
Second, existing RL solutions for LLM agents are not well-suited for multi-turn scenarios.
Inspired by group relative policy optimization (GRPO)~\citep{shao2024deepseekmath}, we extend it to multi-turn settings (M-GRPO) and employ a parallel trajectory rollout strategy to further improve training efficiency by generating multiple trajectories in parallel.
These designs enable efficient RL training and lead to state-of-the-art performance on the WebArena-Lite benchmark, as shown in Figure~\ref{fig:benchmark_overview}.
Extensive ablations further validate our key design choices, reveal an effective test-time scaling strategy for web tasks, and offer insights into the roles of behavior cloning and long CoT reasoning in RL-based web agent training.

Our contributions are summarized as follows:
\begin{itemize}
\item We implement an end-to-end multi-turn RL framework for training web agents, with dynamic context compression and parallel trajectory rollout mechanisms to achieve training efficiency.
\item Based on the proposed M-GRPO algorithm, our method substantially improves task success rates of web agents—boosting Qwen-2.5-3B from 6.1\% to 33.9\% and Llama-3.1-8B from 8.5\% to 44.8\%—surpassing previous state-of-the-art results on the WebArena-Lite benchmark.
\item Extensive analyses and ablation studies underscore the crucial role of behavior cloning, validate the effectiveness of thinking-based prompting and test-time scaling strategies, and provide actionable insights on incorporating long-CoT reasoning in web agents.
\end{itemize}

\section{WebAgent-R1}

\begin{figure*}[!t]
% \vspace{-2em}
\includegraphics[width=\textwidth]{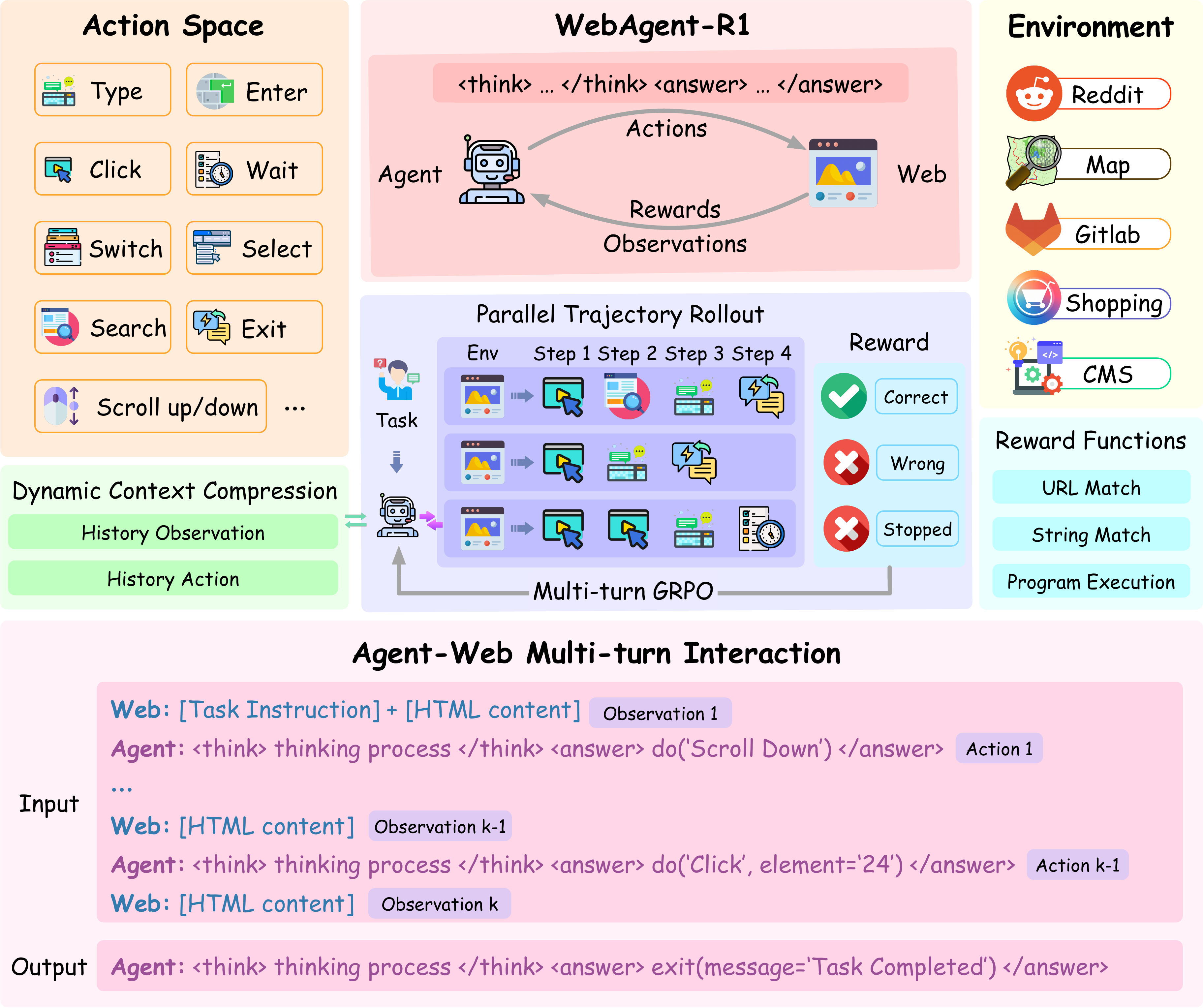}
\vspace{-2em}
\caption{({\bf Top}): Overview of the end-to-end multi-turn RL training framework used in \model. ({\bf Bottom}): An input/output example of agent–web interaction at the $k$-th step. The interaction continues until either the maximum number of steps is reached or the agent generates an \texttt{exit()} action to signal task completion.}
\label{fig:overview}
\vspace{-1.5em}
\end{figure*}

\subsection{Problem Formulation}

We formulate the web task as a Partially Observable Markov Decision Process (POMDP), defined by the tuple $(\mathcal{S}, \mathcal{A}, \mathcal{T}, \mathcal{R})$. At each time step $t$, the agent first observes a state $s_t \in \mathcal{S}$ from the environment $\mathcal{E}$, represented as the text-only HTML content of the current web page. 
Then, it generates an action $a_t$ from a predefined action space $\mathcal{A}$, which includes commonly used web operations.
The environment dynamics $\mathcal{T}(s_{t+1}|s_t,a_t)$ represent how the web page changes in response to actions.
The agent interacts with the environment until either the task is successfully completed or the maximum number of steps is reached.
At the end, the agent receives a binary outcome reward $r_t \in \{0, 1\}$ from reward functions $\mathcal{R}$.

Following prior work~\citep{qi2025webrl}, we adopt WebArena~\citep{zhou2024webarena} as the web environment over other simulated or static environments such as WebShop~\citep{yao2022webshop} or Mind2Web~\citep{deng2023mind2web} for greater practicality---It provides a realistic, self-hostable environment for web agents, along with rule-based rubrics that automatically check for indicators of success in the final state (\eg, confirmation messages or expected content on the page).
Note that some prior works~\citep{liu2025visualagentbench,he2024webvoyager} incorporate web page screenshots as additional visual inputs, whereas our work focuses solely on text-based decision-making over HTML. Other efforts, such as~\citet{yang2025agentoccam}, explore optimizing the action space or prompt design without model fine-tuning. These directions are orthogonal to our investigated problem and can be conceptually integrated with our method as future work.

\subsection{Behavior Cloning}

To initialize the web agent, we first apply behavior cloning (BC) using a fixed dataset of expert demonstrations $\mathcal{D} = \{(h_t, a_t)\}$, where $h_t$ denotes the full interaction history up to time step $t$, defined as $h_t = (s_1, a_1, s_2, a_2, \ldots, s_t)$. The policy $\pi_\theta$ is trained via supervised fine-tuning (SFT) to imitate expert actions conditioned on this history:
\[
\mathcal{L}_{\text{BC}} = - \mathbb{E}_{(h_t, a_t) \sim \mathcal{D}} \left[ \log \pi_\theta(a_t \mid h_t) \right]
\]
This warm-up stage enables the agent to acquire basic web interaction skills defined in the action space. As indicated in our ablation study (\S~\ref{sec:ablation_study}), this BC-trained policy provides a crucial foundation for subsequent reinforcement learning optimization.

\begin{table*}[t!]
\centering
\small
\caption{
Comparison of different methods for training web agents. \textit{Trial-and-Error} indicates whether the method supports learning through interactions with the environment (\ie, reinforcement learning).
\textit{On-Policy} denotes whether the training data is collected from the current policy.
\textit{Replay Buffer Free} indicates methods that do not require selectively sampling trajectories from a replay buffer, a complexity common in off-policy RL. 
\textit{Self-Sufficient} means no external training signals required (\eg, WebRL trains an additional outcome reward model to label new data generated by GPT-4).
As shown, our method is the only one that enables end-to-end RL with on-policy updates while avoiding additional complexities such as maintaining a replay buffer and being free from external supervision.
}
\begin{tabular}{lcccc}
\toprule
Method & Trial-and-Error & On-Policy
& Replay Buffer Free & Self-Sufficient\\
\midrule
Behavior Cloning (SFT)     & {\color{red!60!black}\xmark} & {\color{red!60!black}\xmark} & {\color{green!60!black}\cmark} & {\color{green!60!black}\cmark} \\
AWR~\citep{peng2019advantage}     & {\color{red!60!black}\xmark} & {\color{red!60!black}\xmark} & {\color{red!60!black}\xmark}  & {\color{green!60!black}\cmark}  \\
DigiRL~\citep{bai2024digirl} & {\color{green!60!black}\cmark}  & {\color{red!60!black}\xmark} & {\color{red!60!black}\xmark} & {\color{green!60!black}\cmark} \\
WebRL~\cite{qi2025webrl}   & {\color{green!60!black}\cmark} & {\color{red!60!black}\xmark} & {\color{red!60!black}\xmark} & {\color{red!60!black}\xmark} \\
\textbf{\model} & {\color{green!60!black}\cmark} & {\color{green!60!black}\cmark} & {\color{green!60!black}\cmark}  & {\color{green!60!black}\cmark}  \\
\bottomrule
\end{tabular}
\end{table*}

\subsection{End-to-End Multi-Turn Reinforcement Learning}\label{sec:e2e_rl}

As illustrated in Figure~\ref{fig:overview}, our end-to-end multi-turn RL framework trains web agents through online interactions guided by rule-based outcome rewards.
To enable efficient and scalable training, we implemented two key mechanisms: \emph{dynamic context compression} to reduce memory overhead, and \emph{parallel trajectory rollout} to improve sampling efficiency.
Based on the BC-trained policy, we further fine-tune the agent using an extension of  GRPO~\citep{qi2025webrl} in the multi-turn settings, termed 
\emph{M-GRPO}.
Our implementation can be viewed as a minimalist approach that supports efficient multi-turn RL training while maintaining generality, with potential for future extensions (\eg, allowing fine-grained reward shaping for intermediate steps).

\paragraph{Dynamic Context Compression}
In web tasks, each observation $s_t$ often contains thousands of tokens.
Across multi-turn interactions, the accumulated context grows rapidly, leading to excessive memory usage and potential out-of-memory issues, making training impractical.
To address this, we propose a dynamic context compression strategy. As new observations arrive, earlier ones are simplified to reduce the context length while preserving the complete action history.
Let the interaction history at step $t$ be $h_t = (s'_1, a_1, s'_2, a_2, \ldots, s_t)$, where each $s'_i$ is a template with only a few tokens (\eg, $s'_i=\texttt{``Simplified HTML''}$) representing prior observations.
When the agent executes an action $a_t$ and receives a new observation $s_{t+1}$, the updated history becomes $h_{t+1} = (s'_1, a_1, s'_2, a_2, \ldots, s'_t, a_t, s_{t+1})$, where $s_t$ is replaced by its simplified version $s'_t$.
This allows the agent to maintain a compact yet informative context of past interactions.
Since the context evolves dynamically, we also update the loss masks accordingly to ensure that the loss is correctly computed only on the action tokens during the M-GRPO optimization. 

\paragraph{Multi-turn GRPO}
Inspired by GRPO, we extend its standard form to multi-turn RL settings and introduce multi-turn group relative policy optimization (M-GRPO).
Specifically, for each task $q$, we first sample a group of trajectories $\{\tau_1, \tau_2, \cdots, \tau_G\}$ and then optimize the policy model $\pi_\theta$ by minimizing the following loss:

\begin{scriptsize}    
\begin{align*}
\mathcal{L}_{\text{M-GRPO}}(\theta) = 
- \frac{1}{G} \sum_{i=1}^G 
{\color{black}\frac{1}{|\tau_i|} \sum_{j=1}^{|\tau_i|}} \left(
\frac{1}{|a_{i,j}|} \sum_{t=1}^{|a_{i,j}|} \left[
\tilde{A}_{i,j,t} - \beta\, \mathbb{D}_{\text{KL}}(\theta) 
\right] \right)
\end{align*}
\end{scriptsize}

\noindent where $\tau_i=\{a_{i,1}, a_{i, 2}, \cdots, a_{i,\vert \tau_i\vert}\}$ is the sequence of generated actions in the $i$-th trajectory, {$\tilde{A}_{i,j,t} = \min\{ r_{i,j,t}(\theta) A_{i,j}, \text{clip}(r_{i,j,t}(\theta), 1{-}\epsilon, 1{+}\epsilon) A_{i,j}\}$} is the advantage for the $t$-th token in action $a_{i,j}$ of trajectory $\tau_i$, 
%$r_{i,j,t}(\theta)$
$r_{i,j,t}(\theta) = \frac{\pi_\theta(a_{i,j,t} \mid q, a_{i,j,<t})}{\pi_{\text{old}}(a_{i,j,t} \mid q, a_{i,j,<t})}$
denotes the importance sampling term, $\epsilon$ and $\beta$ are hyper-parameters, and $A_{i,j} = \frac{r_i - {\mathrm mean(\bm{r})}}{{\mathrm std(\bm{r})}}$ is the group relative advantage, computed using a group of rewards $\bm{r}=\{r_1, r_2, \ldots, r_G\}$ produced by rule-based reward functions.

\begin{table*}[!t]
\centering
\small
\caption{Task success rate (SR) comparison across different methods on various websites in WebArena-Lite~\citep{liu2025visualagentbench, qi2025webrl, zhou2024webarena}. Baseline performance is reported as the higher value between our reproduced results and those reported in the literature~\citep{qi2025webrl}. The best scores are highlighted in bold.\label{tab:main_res}}
\begin{tabular}{lcccccccc}
\toprule    
   Method   & Reddit & GitLab & CMS & Map & Shopping & Average SR \\ 

\midrule
\multicolumn{7}{c}{\emph{Prompting Method}} \\
{\bf General Model} \\
\quad Qwen2.5-3B
&5.3  &13.3  &5.7  &0 &4.4 &6.1  \\
\quad Llama3.1-8B
& 5.3 & 10.0 & 5.7 &15.4 &8.9 &8.5  \\
\quad Qwen2.5-32B
& 10.5 & 20.0 & 20.0 & 19.2 & 17.8 & 16.9 \\
\quad  GPT-4o
& 10.5 & 10.0 & 20.0 &20.0 &11.1 &13.9  \\
\quad  GPT-4o-Turbo
& 10.5 & 16.7 & 14.3 &36.7 &13.3 &17.6  \\
{\bf Reasoning Model} \\
\quad QwQ-32B
& 15.8 & 33.3 & 25.7 & 15.4 & 20.0 & 22.4 \\
\quad OpenAI-o3
& 36.8 & 46.7 & 45.7 & {\bf 38.5} & 33.3 & 39.4 \\
\quad OpenAI-o4-mini
& 47.4 & 43.3 & 45.7 & 26.9 & 28.9 & 36.9 \\
\midrule
\multicolumn{7}{c}{\emph{Finetuning Method}} \\

{\bf Qwen2.5-3B} \\
\quad {Behavior Cloning}
& 42.1 & 16.7 & 22.9 & 26.9 & 11.1 & 20.0\\
\quad {\model}
& 26.3 & 53.3 & 48.6 & 26.9 & 24.4 & 33.9 \\

{\bf Llama3.1-8B} \\
\quad {Behavior Cloning}
& 36.8 & 6.7 & 20.0 &33.3 &17.8 &20.6  \\
\quad {Filtered BC}~\citep{pan2024autonomous}
& 52.6 & 20.0 & 31.4 &23.3 &8.9 &23.0   \\
\quad  {AWR}~\citep{peng2019advantage}
& 57.9 & 26.7 & 31.4 &26.7 &17.8 &28.5  \\
\quad {DigiRL}~\citep{bai2024digirl}
& 57.9 & 26.7 & 37.1 &33.3 &17.8 &30.3   \\
\quad  {WebRL}~\cite{qi2025webrl}
& {\bf 63.2} & 46.7 & 54.3 &36.7 &31.1 &42.4   \\
\quad {\model}
& 47.4 & {\bf 56.7} & {\bf 57.1} & 23.1 & {\bf 44.4} & {\bf 44.8} \\

\bottomrule
\end{tabular}
% \vspace{-1.5em}
\end{table*}

\paragraph{Parallel Trajectory Rollout}  
Generating a group of trajectories requires repeated interaction with the environment and can be time-consuming. To address this, we introduce a parallel trajectory rollout strategy, where multiple independent browser instances $\{\mathcal{E}_1, \mathcal{E}_2, \cdots, \mathcal{E}_G\}$ are instantiated, each maintaining its own context (\eg, cookies). For each task, all instances are initialized with the same starting page, but the agent interacts with them independently, resulting in diverse histories and trajectories. This parallel design enables efficient trajectory generation in M-GRPO.

\paragraph{Reward Design}  
We use the default rule-based reward functions in the web environment, which assign binary rewards ($r{=}1$ for success, $r{=}0$ otherwise) based on task-specific criteria (\eg, reaching a target page). This eliminates the need for outcome reward models~\citep{qi2025webrl}, ensuring a simple and generalizable training setup.
\section{Experiments}

\subsection{Experimental Setup}

\paragraph{Web Environment}  
Like prior works~\citep{liu2025visualagentbench,qi2025webrl}, we focus on web agents for real-world scenarios, specifically utilizing WebArena~\citep{zhou2024webarena}, a self-hostable and realistic web environment that supports practical tasks across diverse domains: social forums (Reddit), collaborative coding (GitLab), e-commerce content management systems (CMS), open street maps (Map), and online shopping (Shopping).

\paragraph{Dataset and Evaluation Metrics}  
Following~\citet{qi2025webrl}, we use the public 9,460 trajectories for behavior cloning, and adopt WebArena-Lite, a human-verified version of WebArena, for more reliable evaluation. Specifically, we use 165 verified tasks for evaluation and 647 remaining tasks for RL training. Task success rate is calculated using the built-in rule-based rubrics.

\paragraph{Baselines} For prompting baselines, we provide a comprehensive comparison with both open-source and proprietary models, including general-purpose models (\eg, Qwen2.5, Llama3.1, GPT-4) and reasoning-specialized models (\eg, QwQ, OpenAI o3~\citep{openai2025o3}), covering various model sizes. For finetuning methods, we employ Qwen2.5-3B and Llama3.1-8B as the backbone model. 
We refer the readers to~\citep{liu2025visualagentbench} for more baseline results on the WebArena-Lite benchmark.

\paragraph{} More details on the environment and implementation are provided in Appendix~\ref{appendix:web_env} and~\ref{appendix:imple}.
We also provide the prompt templates and qualitative examples in Appendix~\ref{appendix:prompt} and~\ref{appendix:demo}.

\subsection{Main Results} 

\begin{figure*}[!t]
\vspace{-1em}
\centering
\begin{subfigure}[b]{0.325\textwidth}
    \includegraphics[width=\textwidth]{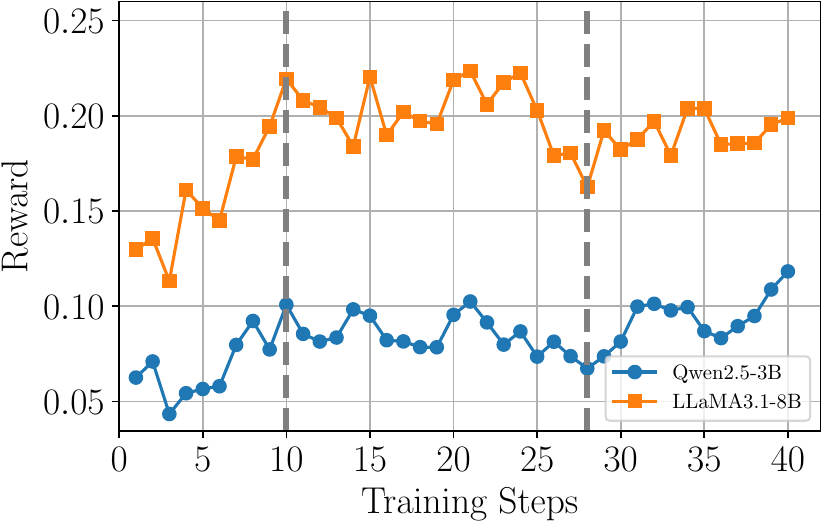}
    \caption{ Reward\label{fig:reward}}
\end{subfigure}
\hfill
\begin{subfigure}[b]{0.325\textwidth}
    \includegraphics[width=\textwidth]{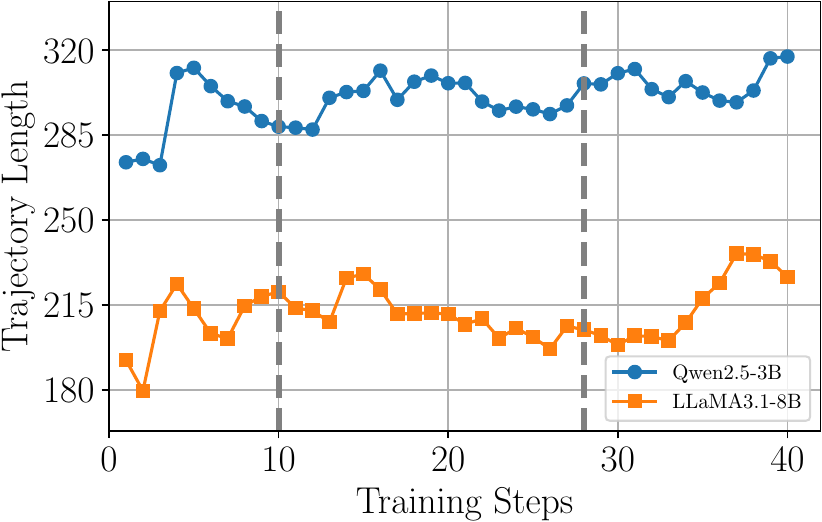}
    \caption{Trajectory Length\label{fig:response}}
\end{subfigure}
\hfill
\begin{subfigure}[b]{0.325\textwidth}
    \includegraphics[width=\textwidth]{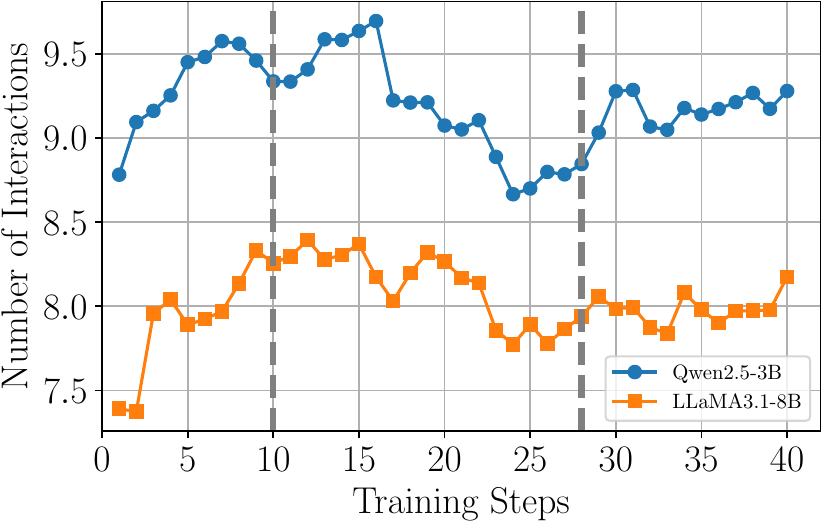}
    \caption{Number of Interactions \label{fig:interaction}}
\end{subfigure}
\caption{Training dynamics during RL, including rewards, trajectory length, and number of interactions. As indicated by the dashed vertical lines in the figure, the entire process can be broadly divided into three phases: (1) initial skill acquisition, (2) exploration for policy refinement, and (3) final policy stabilization.}\label{fig:rl_dynamics}
\end{figure*}

\paragraph{Most LLMs still struggle with web tasks through prompting, highlighting the importance of finetuning for web agents.}  
As shown in Table~\ref{tab:main_res}, our experiments reveal the limitations of off-the-shelf models in web tasks. Despite their strong general capabilities, state-of-the-art models such as OpenAI’s o3 achieve only a 39.4\% success rate (SR).  
In contrast, a finetuned 3B model trained with simple behavior cloning achieves a success rate of 20\%, outperforming proprietary models like GPT-4o.  
We speculate that the poor performance of off-the-shelf models is not due to base model size or capability, but rather to insufficient understanding of HTML structure and web-specific behaviors, as evidenced by the observation that both 3B and 8B models achieve comparable performance after behavior cloning.
These findings emphasize the necessity of domain-specific training on web data to develop effective LLM-based web agents.

\paragraph{Reasoning models are better web agents.}  
Compared to general-purpose LLMs, models equipped with explicit thinking capabilities perform significantly better on web tasks, likely due to their ability to decompose high-level goals and explicitly lay out dynamic changes in the web interface. This gap underscores the importance of thinking in web environments, which typically require multi-turn decision-making and dynamic contextual understanding.  
Motivated by this observation, we further explore the integration of thinking mechanisms into web agents through prompt design (\S~\ref{sec:analysis}) and training strategies (\S~\ref{sec:ablation_study}), which further confirms the advantage of thinking ability for web agents.

\paragraph{Reinforcement learning enables stronger performance for web agents.}
While behavior cloning via SFT can significantly improve LLM's performance as web agents (\eg, boosting Qwen2.5-3B from 6.1\% to 20\%), applying RL on top of the SFT-trained policy leads to additional substantial gains (\eg, further boosting Qwen2.5-3B from 20\% to 33.9\%).
We attribute these improvements to RL’s ability to optimize long-horizon decision-making, explore novel strategies beyond those seen in the SFT data through trial-and-error across dynamic web interactions.
While prior RL solutions for web agents, such as DigiRL and WebRL, have also shown performance gains, our method achieves even stronger results, highlighting the effectiveness of our end-to-end multi-turn RL framework.

\subsection{Training Dynamics }

\begin{figure*}[!t]
\centering
\begin{subfigure}[b]{0.325\textwidth}
    \includegraphics[width=\textwidth]{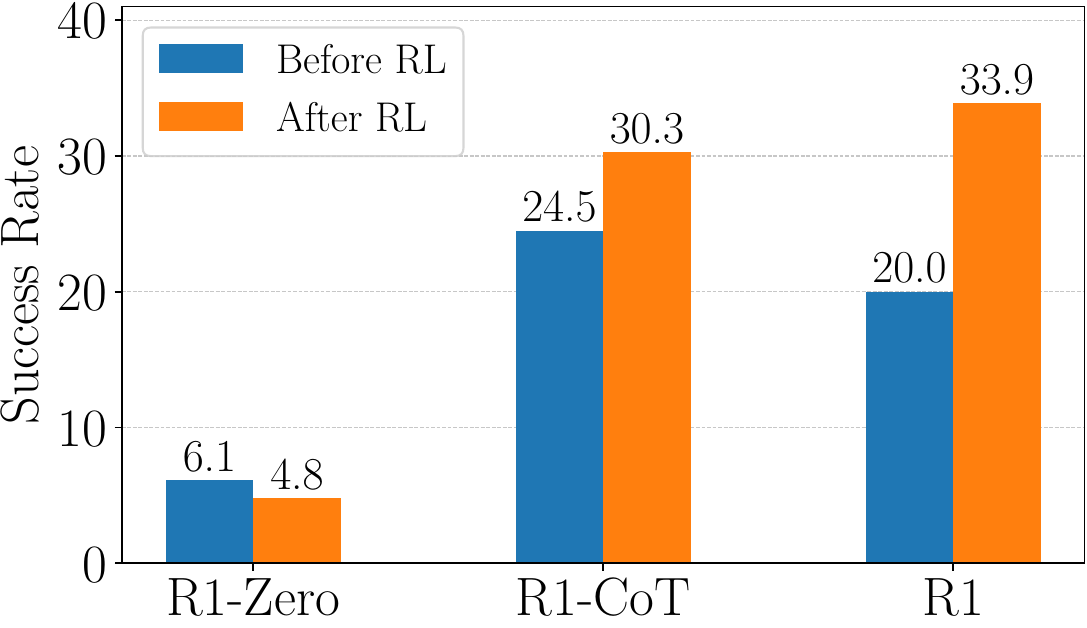}
    \caption{ Success Rate\label{fig:ablation_SR}}
\end{subfigure}
\hfill
\begin{subfigure}[b]{0.325\textwidth}
    \includegraphics[width=\textwidth]{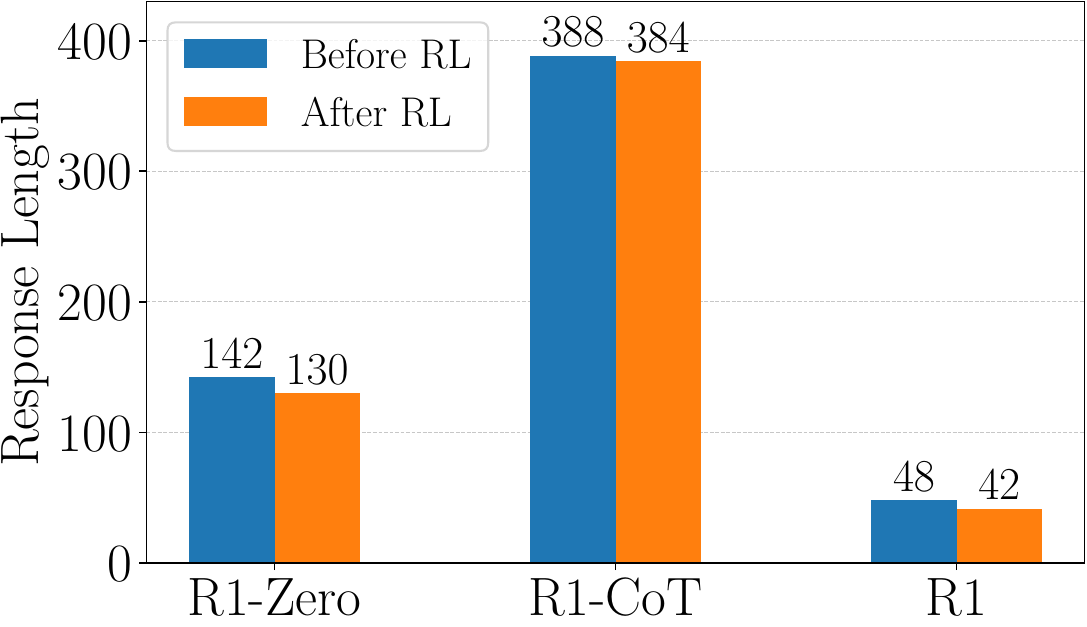}
    \caption{Response Length\label{fig:ablation_res_len}}
\end{subfigure}
\hfill
\begin{subfigure}[b]{0.325\textwidth}
    \includegraphics[width=\textwidth]{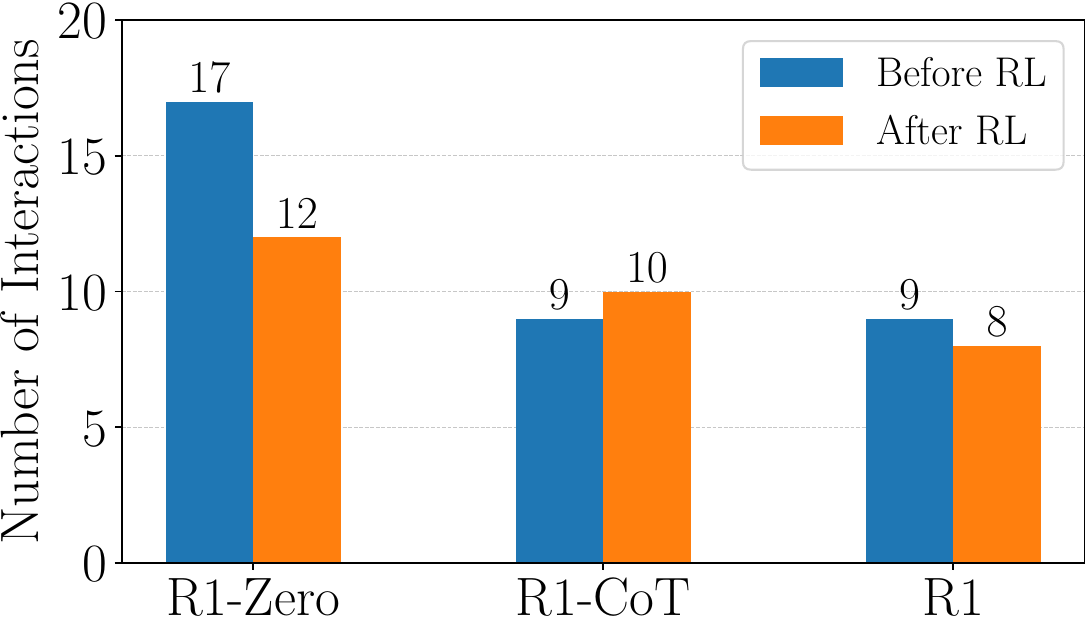}
    \caption{Number of Interactions \label{fig:ablation_n_rounds}}
\end{subfigure}
\vspace{-0.5em}
\caption{
Ablation study on RL initialization policy by comparing \model{} (R1) with two variants: \modelzero{} (R1-Zero), initialized from an off-the-shelf model without SFT, and \modelcot{} (R1-CoT), initialized from an SFT model trained with long chain-of-thought (CoT) data during behavior cloning. The comparison includes task success rate, single-turn response length, and number of interactions, evaluated both before and after applying RL.}
\label{fig:ablation}
\vspace{-1em}
\end{figure*}

To understand how the proposed end-to-end reinforcement learning optimizes the behavior of the web agents, we analyze the training dynamics across three metrics: reward, trajectory length (i.e., the number of tokens in model responses across all multi-turn interactions), and number of interactions. As shown in Figure~\ref{fig:rl_dynamics}, the learning process can be broadly divided into three distinct phases, separated by vertical dashed lines.

\paragraph{Reward.}
Phase 1 shows a rapid increase in reward, indicating that the agent quickly learns basic skills and begins to succeed on simpler tasks. In Phase 2, the reward growth plateaus and slightly fluctuates, suggesting that the agent is exploring different strategies and refining its policy. In Phase 3, reward gradually improves again, indicating exploitation and increased stability.

\paragraph{Trajectory Length.}
Trajectory length increases sharply during Phase 1, then stabilizes in Phase 2. In Phase 3, a modest increase is observed again. This trend suggests that the agent initially learns to produce more detailed outputs, followed by a period of consolidation and later refinement to balance verbosity with task effectiveness.

\paragraph{Number of Interactions.}
The number of interaction rounds increases during Phase 1 as the agent becomes more proactive, followed by a reduction in Phase 2 as it learns to interact more efficiently. In Phase 3, the interaction count stabilizes, indicating convergence toward a more consistent and effective interaction strategy.

\paragraph{} These trends highlight a three-phase learning dynamic commonly observed in RL: (1) initial skill acquisition, (2) exploration for policy refinement, and (3) final policy stabilization.
Interestingly, both Qwen2.5-3B and Llama3.1-8B follow similar learning patterns, suggesting that our end-to-end multi-turn RL framework effectively scales across model sizes and enables stable policy improvement.

\subsection{Ablation Study}\label{sec:ablation_study}

To validate key design choices in our framework, we conduct a set of ablation studies using Qwen2.5-3B as the backbone model. 
Specifically, we introduce two variants, \modelzero{} and \modelcot{}, to study the impact of behavior cloning and long CoT for web agents.
The results are presented in Figure~\ref{fig:ablation}.

\paragraph{Behavior cloning is crucial for training web agents with RL.} 
\modelzero{} skips the behavior cloning stage and starts RL directly from an off-the-shelf model, with an initial success rate of only 6.1\%. Surprisingly, the model’s performance even deteriorates slightly after RL.
We hypothesize that this is due to the lack of knowledge about web tasks since the model tends to produce incomplete or ill-formed actions (\eg, missing required arguments) and rarely obtains positive rewards during RL. This severely hampers effective exploration and learning, highlighting that behavior cloning is essential for initializing web agents and enabling successful subsequent RL.

\paragraph{Incorporating long-CoT data into behavior cloning leads to more performant web agents.}
We first augment the behavior cloning (BC) data by generating long-CoT traces using a strong reasoning model (see Appendix~\ref{appendix:data_aug} for details), and then apply SFT to obtain a \emph{long-CoT SFT} model (\ie, the \modelcot{} variant before RL).
Compared to the SFT model trained on standard BC data, the long-CoT SFT model achieves a much higher task success rate (24.5\% vs. 20\%), demonstrating the effectiveness of long-CoT reasoning for web agents.

\paragraph{Limited gains from RL for long-CoT SFT model.}
While RL shows promising improvements for both the vanilla SFT and long-CoT SFT models, it is interesting that the gain is notably smaller for the latter. Specifically, \model{} improves from 20\% to 33.9\%, whereas \modelcot{} improves from 24.5\% to only 30.3\%.
We hypothesize that this is because the deterministic reasoning patterns learned during long-CoT BC may constrain the model’s exploration space during RL, limiting its ability to discover novel strategies compared to standard SFT models with more flexible exploratory behaviors.

\subsection{Analysis}\label{sec:analysis}

\begin{table}[!t]
\centering
\footnotesize
\caption{Analysis of prompting design. We report the average success rate (SR), single-turn response length, and number of interactions. The result reveals a novel test-time scaling paradigm by increasing the number of interactions for multi-turn interactive web tasks.
\label{tab:prompt_design}}
\vspace{-.5em}
\resizebox{0.485\textwidth}{!}{
\begin{tabular}{lccc}
\toprule
Method & SR & Length & \# of Interactions  \\
\midrule
W/o thinking format  \\
\quad Qwen2.5-3B
& 3.2  & 139 & 6   \\
\quad Llama3.1-8B
& 4.8   & 43  & 7   \\
\quad o4-mini
& 15.9   & 56  & 5   \\

With thinking format \\
\quad Qwen2.5-3B
& 6.1  & 142 & 17    \\
\quad Llama3.1-8B
& 8.5   & 39  & 11   \\
\quad o4-mini
& 36.9  & 57 & 10    \\

\bottomrule
\end{tabular}
}
\vspace{-1em}
\end{table}

\begin{figure}[!t]
\centering
\includegraphics[width=0.425\textwidth]
{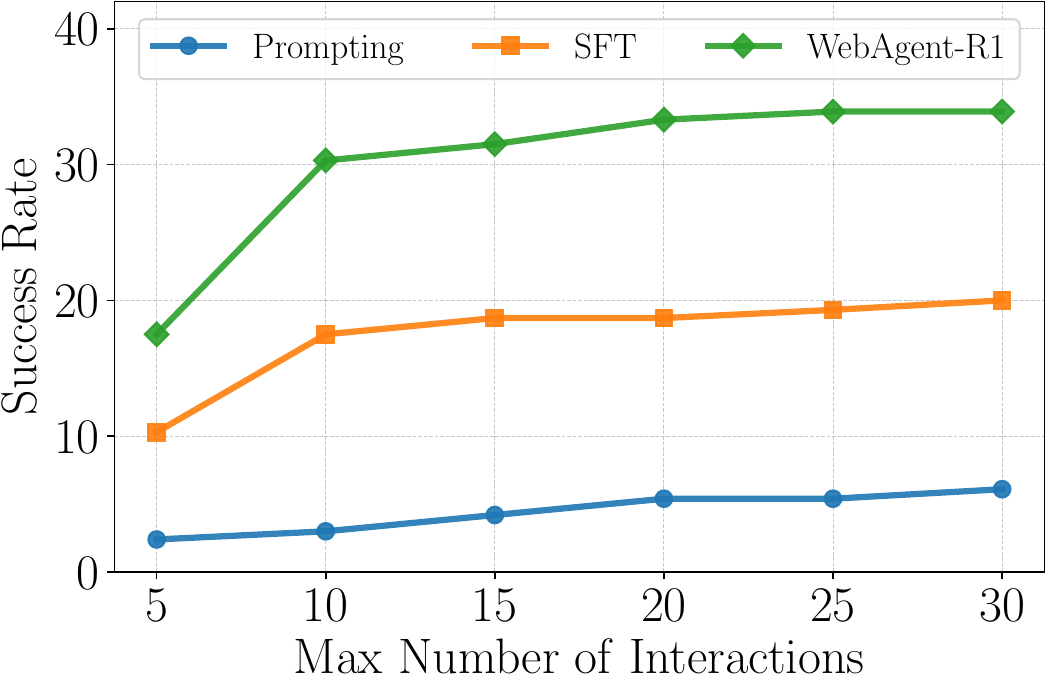}
\caption{Analysis of test-time scaling with increased max number of interactions. Allowing more interactions enables the web agent to produce longer trajectories and consistently improves the success rate.}
\label{fig:scaling_interaction}
\vspace{-1em}
\end{figure}

\paragraph{Prompting with thinking format unleashes the potential of LLMs as web agents.}
As shown in Table~\ref{tab:prompt_design}, using the thinking format significantly improves task success rates across models, particularly for stronger ones (\eg, o4-mini improves from 15.9\% to 36.9\%).
Interestingly, while the average single-turn response length remains similar (\eg, 139 $\rightarrow$ 142 tokens for Qwen2.5-3B), the number of interactions increases substantially (\eg, 6 $\rightarrow$ 17 for Qwen2.5-3B) with the thinking format.
These results indicate that prompting with explicit thinking instructions enhances web agents by encouraging more frequent interactions.
As a result, we conjecture that this observation suggests a novel test-time scaling strategy for web tasks---rather than producing longer single-turn responses, the web agent can become more effective by engaging in deeper multi-turn interactions.

\paragraph{Test-time scaling through increased interactions leads to better performance on web tasks.}
Building on the above finding, we further investigate how increasing the number of interactions between the web agent and the environment affects performance. As shown in Figure~\ref{fig:scaling_interaction}, allowing more interaction turns consistently improves success rates across prompting-based, SFT (\ie, behavior cloning), and RL-based methods.
We hypothesize that this form of test-time scaling facilitates deeper exploration and yields longer trajectories, potentially enabling the agent to iteratively refine its actions and make more informed decisions through extended interactions.

\paragraph{\model{} generalizes well to out-of-distribution (OOD) tasks.}
We conducted additional evaluation on the WebVoyager benchmark using Qwen2.5-3B as the baseline model.
The benchmark covers diverse domains that are unseen in the WebArena environment and thus serves as an OOD evaluation of our method. 
Specifically, we randomly sample 25 tasks for each of 5 domains in WebVoyager as the OOD evaluation set to compare our method with both the prompting baseline and the SFT variant without any further training.
As shown in Table~\ref{tab:ood_eval}, \model{} consistently outperforms both prompting and SFT baselines across all domains, confirming the effectiveness and generalizability of our method.

\begin{table}[!t]
\centering
\footnotesize
\caption{Out-of-distribution (OOD) evaluation on five domains from the WebVoyager benchmark. We report the success rates of compared methods on each domain.\label{tab:ood_eval}}
\vspace{-.5em}
\resizebox{0.485\textwidth}{!}{
\begin{tabular}{lccc}
\toprule
Domain & Prompting & SFT & \model{}  \\
\midrule

Allrecipes
& 0\%  & 4\% & 28\%    \\
Amazon
& 4\%   & 4\%  & 24\%   \\
Arxiv
& 20\%   & 20\%  & 24\%   \\
Coursera
& 16\%  & 16\% & 44\%    \\
Google Map
& 4\%   & 16\%  & 40\%   \\
Average
& 8.8\% & 12\%  & 32\%   \\

\bottomrule
\end{tabular}
}
\vspace{-1em}
\end{table}
\section{Related Works}

\subsection{LLM-based Agents}

LLMs have demonstrated promising agentic capabilities, such as breaking down complex tasks into manageable subgoals and reasoning over long horizons~\citep{zhou2022show,huang2022language,madaan2022language,li2023camel,li2023take,wu2024autogen,liu2025selfelicit,chu2025llm}. Building on these capabilities, LLM-based agents have been applied to a variety of real-world interactive tasks, including web navigation~\citep{nakano2021webgpt,yao2022webshop,ma2023laser,gur2024a,abuelsaad2024agent,lutz2024wilbur,patel2024large,putta2024agent}, general computer use~\citep{li2020mapping,deng2023mind2web,yang2024swe}, and embodied environments~\citep{puig2018virtualhome,shridhar2020alfred,toyama2021androidenv,fan2022minedojo,huang2022language}.
Specifically, our work focuses on text-based web agents that operate in browser-based environments purely based on HTML content, which requires agentic capabilities such as tool use, memory, and decision-making under partial observability~\citep{zhou2024webarena,qi2025webrl}.
Complementary to this line of work, GUI agents leverage additional multimodal inputs such as screenshots, enabling visual-guided interactions with the environment~\citep{lee2023pix2struct,shaw2023pixels,zheng2024gptvision,he2024webvoyager,he2024openwebvoyager,koh2024visualwebarena,kil2024dual,lei2025scaffolding,liu2025visualagentbench}.
For a comprehensive overview, we refer readers to recent surveys~\citep{wang2024survey,tseng2024two,hu2025agents,ning2025survey}.

\subsection{Reinforcement Learning for LLMs}

Recent advances like DeepSeek-R1~\citep{guo2025deepseek} highlight the strong potential of RL in enhancing LLMs. However, most prior work focuses on single-turn tasks such as math problems~\citep{shao2024deepseekmath, zhu2025surprising, shao2025spurious, ouyang2025rast,wei2025truthrl}, with limited exploration in multi-turn settings~\citep{zhou2024archer,zhou2025sweet}.
Recent efforts have made some progress in this direction, such as training LLM agents to repeatedly use search engines~\citep{jin2025search, sun2025zerosearch, chen2025research, song2025r1,wang2025outcomerewarddecouplingsearch}, but typically constrain actions to simple API calls without real environment interaction.
A few concurrent works, such as RAGEN~\citep{wang2025ragen} and SkyRL~\citep{cao2025skyrl}, have applied RL to more dynamic settings like simulated games and coding environments~\citep{jimenez2024swebench}. However, real-world web environments remain largely underexplored.
Our work fills this gap by providing a practical framework and offering actionable insights for training web agents with end-to-end RL.
\section{Conclusion}
This work introduces \model, an end-to-end multi-turn RL framework for training web agents.
We extend the standard GRPO to multi-turn settings, termed M-GRPO, and implement dynamic context compression and parallel trajectory rollout mechanisms for efficient training.
Empirically, \model achieves new state-of-the-art results on the WebArena-Lite benchmark.
Our findings underscore the critical role of behavior cloning in initializing web agents, providing a strong foundation for effective RL.
We further analyze training dynamics and explore the effects of thinking-based prompting and test-time scaling strategies, showing that increasing interaction depth consistently enhances web agents.
Future work includes exploring multi-modal inputs and extending our approach to broader GUI-based tasks beyond web environments, such as computer use.

\section*{Limitations and Potential Risks}
Despite the effectiveness of \model, our current approach has several limitations that suggest directions for future work.
First, we consider only textual input for the web tasks. Incorporating additional visual input (\eg, screenshots) may enhance performance since visual information, such as layout and colors, can be helpful for effective navigation and decision-making.
Second, our method relies on rule-based outcome rewards to guide RL training. While effective in our setting, such reward functions may not be readily available in other interactive scenarios, such as open-ended travel planner agents, where task goals are ambiguous and no clear reference or verifiable outcome is available.
Lastly, like existing web agents, our model is trained with a fixed set of predefined actions (\eg, click, type), which can limit its flexibility when encountering interactive elements that require unseen operations. Enabling dynamic adaptation to new operations remains an open challenge for web agents. 

In terms of potential risks, such agents should be used with caution when deployed in real-world environments, especially those involving administrative privileges. For example, when interacting with content management systems (CMS) in a production environment, the agent may inadvertently perform destructive actions, such as modifying or deleting sensitive data. To ensure safe deployment, future work should incorporate permission controls, verification prompts, and safeguards to prevent high-impact or irreversible actions.

\section*{Acknowledgments}
The authors would like to thank Yu Meng and Shiyu Feng from the University of Virginia for their valuable feedback and discussions.
We also thank anonymous reviewers for their constructive and insightful comments.

\bibliography{custom}

\appendix

\section{Web Environment}\label{appendix:web_env}

\paragraph{WebArena-Lite}  
WebArena~\citep{zhou2024webarena} is a realistic, self-hostable web environment for developing LLM-based agents. It comprises 812 real-world web tasks spanning diverse domains, including social forum (Reddit), collaborative coding (GitLab), e-commerce content management system (CMS), open street map (Map), and online shopping (OneStopShop).
WebArena-Lite~\citep{liu2025visualagentbench} is a curated version of WebArena designed for more reliable evaluation. It selects 165 representative tasks for human verification as the evaluation set and uses the remaining 647 tasks for training.
It also provides 9,460 trajectories automatically annotated by program-based solvers for behavior cloning. For each website, the authors~\citep{liu2025visualagentbench} summarize the core functionalities and valid items and construct a set of task prototypes and manually implement rule-based solvers using Playwright scripts for each prototype. The corresponding solvers are executed on the websites to collect ground-truth trajectories.
In total, this produces 1,186 valid training samples comprising 9,460 trajectories, released under the Apache License 2.0.

\paragraph{Action Space}  
Agents interact with the environment through a set of predefined actions, including:
\begin{itemize}[itemsep=2pt, parsep=0pt]
  \item \textbf{Click}: simulates a left mouse click on a webpage element.
  \item \textbf{Right Click}: performs a right-click on a specified element.
  \item \textbf{Type}: inputs a text string into an input field.
  \item \textbf{Search}: enters a search query and triggers a search operation.
  \item \textbf{Hover}: moves the cursor over a specific element to reveal tooltips or hidden menus.
  \item \textbf{Scroll Up / Scroll Down}: scrolls the page vertically.
  \item \textbf{Press Enter}: simulates pressing the Enter key, typically after typing.
  \item \textbf{Switch Tab}: changes the current browser tab.
  \item \textbf{Select Dropdown Option}: selects an option from a dropdown menu.
  \item \textbf{Wait}: pauses the agent’s interaction for a brief period.
  \item \textbf{Exit}: terminates the current session with a final message.
  \item \textbf{Go Backward / Go Forward}: navigates backward or forward in the browser history.
\end{itemize}

\paragraph{Rule-based Metrics}  
In real-world web tasks, there are typically no closed-form solutions, and multiple trajectories may lead to successful task completion. Therefore, we evaluate agents solely based on whether the final goal is achieved and calculate the Success Rate (SR), which indicates whether a task is successfully completed according to the following rule-based evaluation metrics:

\begin{itemize}[itemsep=2pt, parsep=0pt]
\item \textbf{String Match}: The agent must provide an answer string that matches the expected output.
\item \textbf{URL Match}: The agent is required to navigate to a specific webpage. Success is determined by comparing the final URL to a reference URL.
\item \textbf{Program Execution}: The agent must modify webpage content or configuration. Evaluation is performed by executing a rule-based script to extract and verify the final state of the page.
\end{itemize}

Each task in WebArena is associated with one of these evaluation metrics, along with the corresponding reference answer, target URL, or validation script when applicable.
This diverse rule-based metric design ensures consistent evaluation across a wide range of web tasks, while accommodating different task objectives and output formats.

\begin{table*}[ht!]
\centering
\small
\caption{Comparison of model outputs from \model{} and \modelcot{}. We present successful trajectories from both models on the same task (\emph{``What are the top-3 best-selling products in Jan 2023?’’}), showing only the first two steps for clarity (a full trajectory is shown in Figure~\ref{fig:demo_cms} for additional context). Compared to \model{}, the long-CoT variant \modelcot{} exhibits a more detailed thinking process.}\label{tab:output_study}
\vspace{-0.5em}
\renewcommand{\arraystretch}{1.3}
\setlength{\tabcolsep}{8pt}
% \rowcolors{2}{pink!10}{gray!10}
\rowcolors{2}{gray!10}{gray!5}
% \rowcolors{2}{orange!5}{orange!10}
\begin{tabular}{>{\centering\arraybackslash}m{3.2cm} p{11.5cm}}
\rowcolor{gray!30}
% \rowcolor{white}
\textbf{Model} & \parbox{\linewidth}{\centering \textbf{Model Output}} \\

\multirow{2}{*}{
  \centering
  \parbox[c]{3cm}{
    \centering
    \model\\
    (Step 1)
  }
} &
{\raggedright\ttfamily 
<think> \# Element: the `REPORTS' section on the left sidebar. </think>\newline
<answer> \textbf{do(action=``Click", element=``7")} </answer>} \\[2pt]

% \rowcolor{white}
\multirow{3}{*}{
  \centering
  \parbox[c]{3cm}{
    \centering
    \model\\
    (Step 2)
  }
} &
{\raggedright\ttfamily
<think> \# Element: the `Bestsellers' report option under the `Products' section in the `REPORTS' </think>\newline
<answer> \textbf{do(action=``Click'', element=``27'')} </answer>} \\[2pt]

% \rowcolor{gray!10}
\multirow{9}{*}{
  \centering
  \parbox[c]{3cm}{
    \centering
    \modelcot\\
    (Step 1)
  }
} &
{\raggedright\ttfamily 
<think> I need to find the top-3 best-selling products for January 2023. Let me look at the current dashboard. The left sidebar has a "Reports" section (element 7) which is critical for accessing sales data. The dashboard's main content shows a table with a product (``Sprite Yoga Strap 6 foot'') and quantity (6), but this appears to be static local sales data rather than a report. Since the user needs a specific monthly ranking, the Reports section is the logical path to business intelligence tools that can aggregate and sort data by time period. Clicking ``Reports'' will likely reveal options for sales analytics, performance tracking, or product rankings required to identify top-selling items in January 2023. </think>\newline
<answer> \textbf{do(action=``Click'', element=``7'')} </answer>} \\

\multirow{18}{*}{
  \centering
  \parbox[c]{3cm}{
    \centering
    \modelcot\\
    (Step 2)
  }
} &
{\raggedright\ttfamily

<think> Alright, let's break down my thought process here. The user wants the top-3 best-selling products for January 2023. I previously clicked the ``Reports'' menu (element 7) to access analytical sections. Now, analyzing the current HTML structure:
\newline 1. **Locating the Reports Hierarchy**:
\newline The Reports section's dropdown (div data-bbox=``88,92,1072,720'') contains nested menus. Under the ``Products'' subsection (span data-text=``Products''), I see options like ``Views'', **``Bestsellers''**, ``Low Stock'', etc.
\newline 2. **Identifying the Target Element**:
\newline   The element with id=``27'' is explicitly labeled `Bestsellers', which directly matches the user's request. This is the most logical path to access sales performance data.
\newline 3. **Ruling Out Distractors**:
\newline Other options like ``Orders'' or ``Tax'' under the Sales section don't specifically target product performance. The ``Bestsellers'' report is tailored to rank products by sales volume, making it the correct choice.
\newline 4. **Next Step Justification**:
\newline Clicking ``Bestsellers'' (element 27) will likely load a report listing products by sales quantity or revenue. From there, I can extract the top-3 entries for January 2023.
\newline **Decision**: Proceed to click element 27 to access the Bestsellers report, which is the most direct route to fulfill the user's query. </think>
\newline <answer> \textbf{do(action=``Click'', element=``27'')} </answer>} \\

\end{tabular}
\vspace{-0.5em}
\end{table*}

\section{Implementation Details}\label{appendix:imple}
We implement our method using Qwen2.5-3B and Llama3.1-8B as the backbone models. By default, we use the instruction-tuned version for both prompting and fine-tuning methods.  
The reinforcement learning (RL) initialization policy is derived from the supervised fine-tuning (SFT) checkpoint obtained via behavior cloning.
Since WebRL leverages additional GPT-4 generated data to train Llama3.1-8B, we ensure a fair comparison by initializing our RL policy with their publicly released checkpoint and applying our end-to-end RL using only the original 647 training tasks, without introducing any extra data.

Our models are trained on a single node of 8 NVIDIA A100 GPUs with 80GB memory via full-parameter fine-tuning.
To optimize GPU utilization, we adopt DeepSpeed~\citep{rajbhandari2020zero} for distributed training with ZeRO-3 offload, along with gradient checkpointing, FlashAttention-2~\citep{dao2024flashattention}, and bf16 mixed precision training enabled for computation efficiency.
For SFT, we use a learning rate of 5e-5 and a batch size of 128, with a cosine LR scheduler for 5\% warmup steps.
For RL training, we use a constant learning rate of 1e-6 with a batch size of 16. The KL divergence regularization coefficient $\beta$ and the clip ratio $\epsilon$ are set to 0.001 and 0.2, respectively.
The maximum context length and number of new tokens are set to 16,384 and 1024. For efficient LLM rollouts during M-GRPO, we use vLLM~\citep{kwon2023efficient} with a tensor parallel size of 1 and a GPU memory utilization ratio of 0.7. Rollout sampling is performed with both the temperature and top-p set to 1.0.

\section{Data Augmentation for Behavior Cloning with long-CoT Traces}\label{appendix:data_aug}
As introduced in the ablation study (\S~\ref{sec:ablation_study}), we augment the original behavior cloning data by generating long-CoT traces using a strong reasoning model, QwQ-32B. We then apply SFT to obtain a long-CoT SFT model, followed by RL training to obtain \modelcot{}.
 As shown in Table~\ref{tab:output_study}, \modelcot is able to generate more detailed thinking compared to \model. 
 
\section{Prompt Templates}\label{appendix:prompt}
The prompt used for data augmentation is shown in Table~\ref{tab:data_aug_prompt}. We define the action space in the system prompt, which is presented in Table~\ref{tab:sys_prompt}. By default, we use the version with the thinking format.

\section{Qualitative Examples}\label{appendix:demo}

% We present one real-world task example with a successful trajectory generated by \model{} for five websites in WebArena in Figure~\ref{fig:demo_cms, fig:demo_reddit, fig:demo_gitlab, fig:demo_map, fig:demo_shopping}.

\begin{figure*}[!t]
\includegraphics[width=\textwidth]{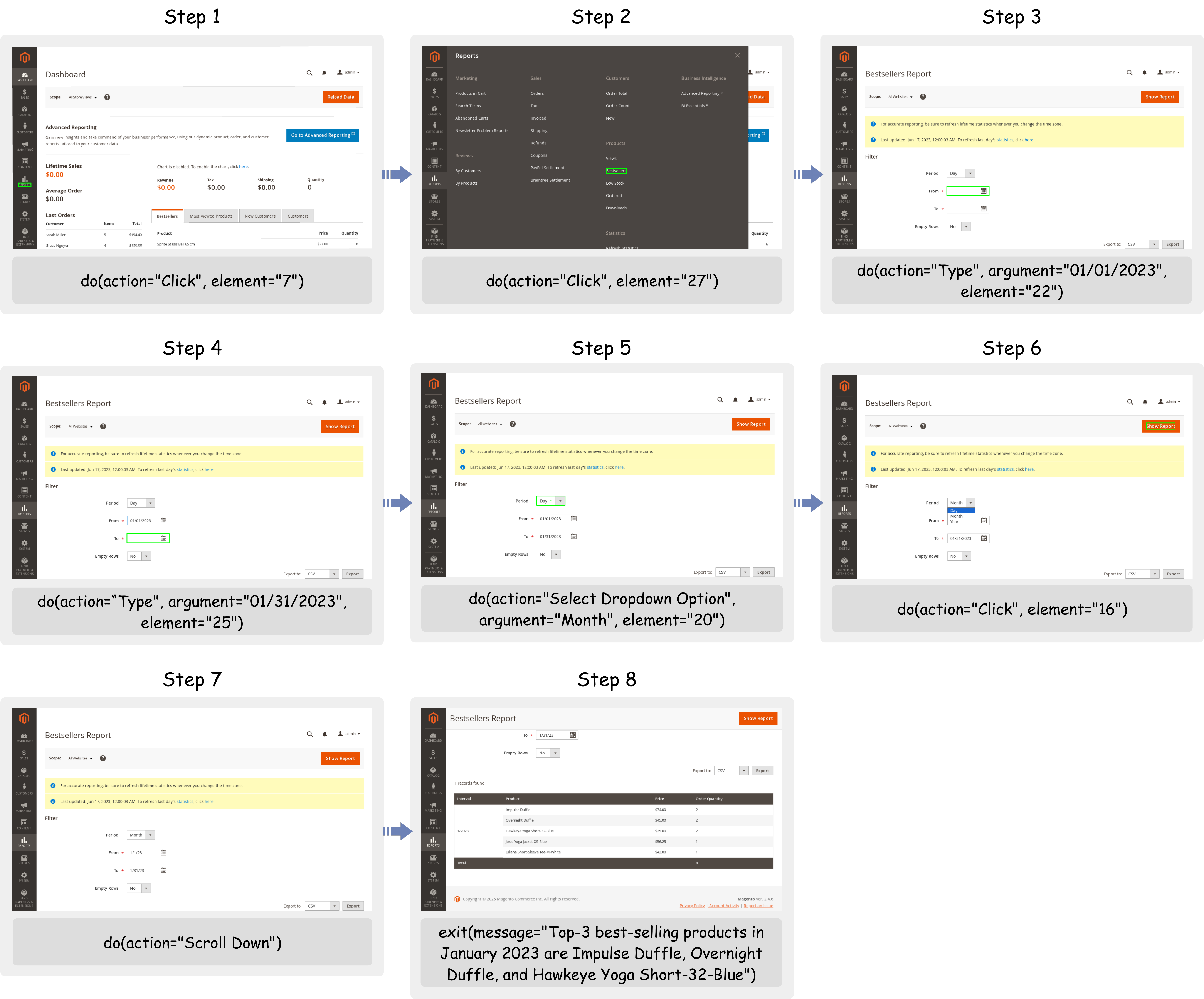}
\caption{A real-world example of a successful trajectory generated by \model on the CMS task: \emph{``What are the top-3 best-selling products in Jan 2023?''}.}
\label{fig:demo_cms}
\end{figure*}

\begin{figure*}[!t]
\includegraphics[width=\textwidth]{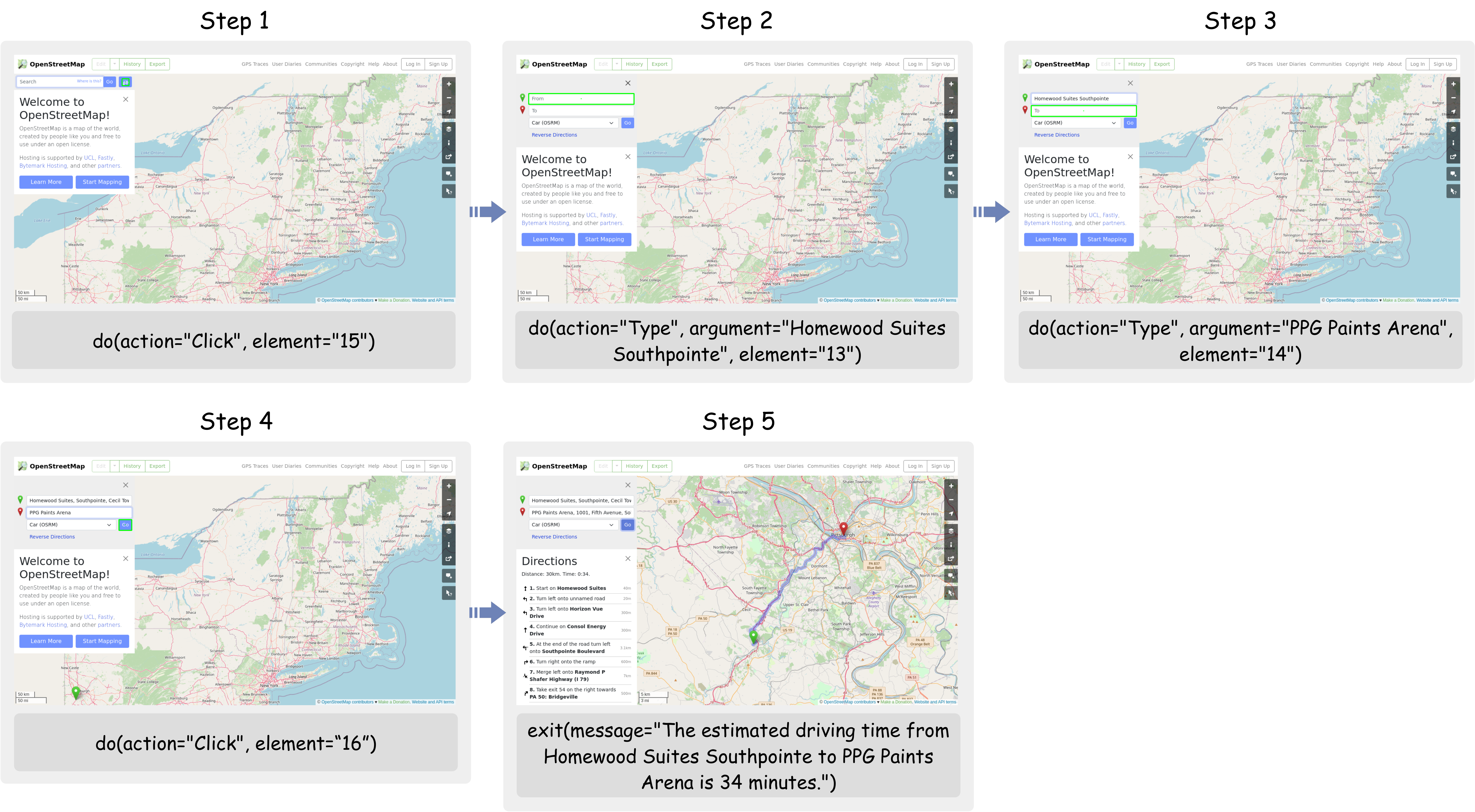}
\caption{A real-world example of a successful trajectory generated by \model on the Map task: \emph{``From my stay at Homewood Suites Southpointe, what's the estimated driving time to reach PPG Paints Arena?''}.}
\label{fig:demo_map}
\end{figure*}

\begin{figure*}[!ht]
\includegraphics[width=\textwidth]{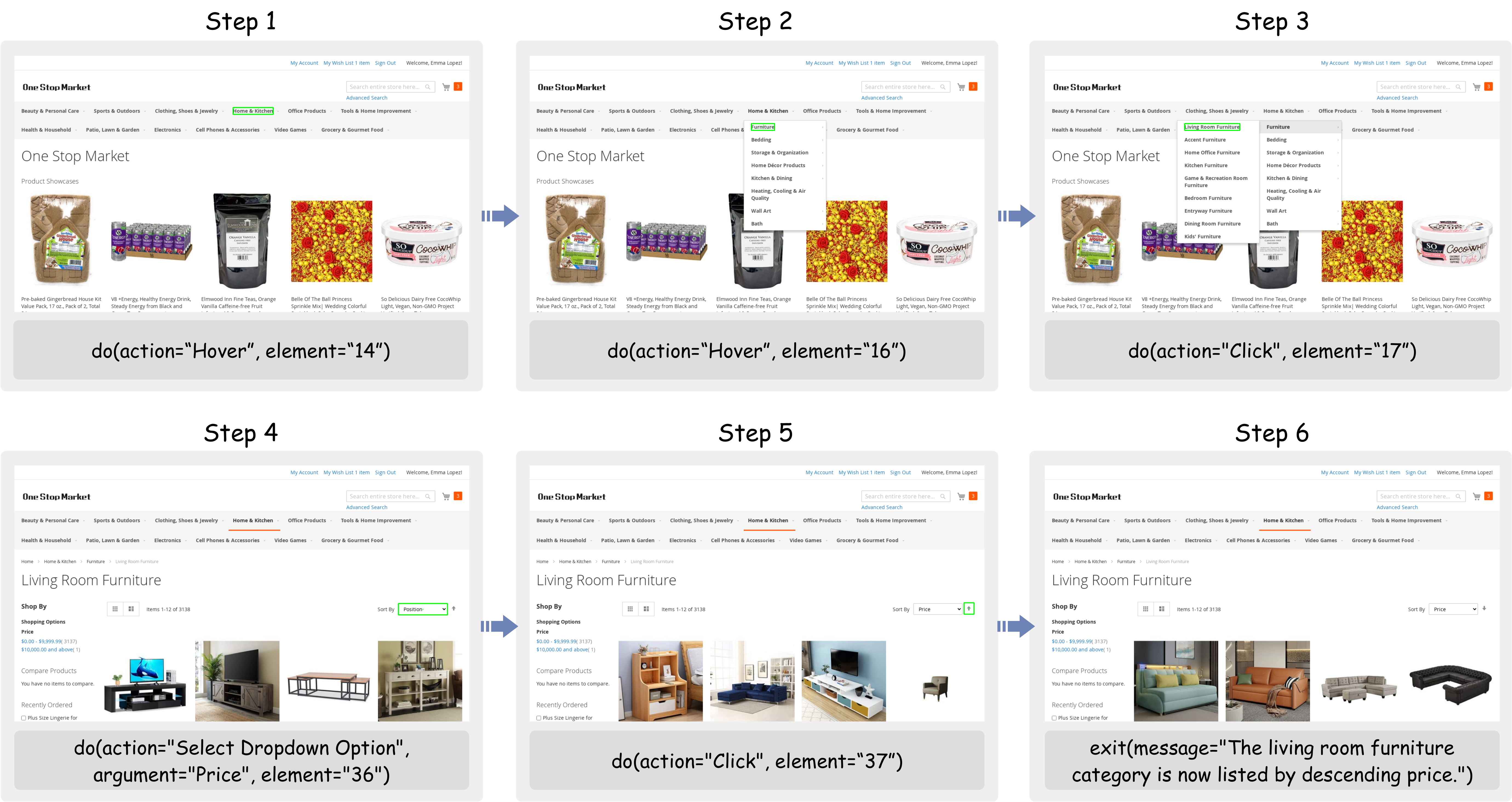}
\caption{A real-world example of a successful trajectory generated by \model on the Shopping task: \emph{``List products from living room furniture category by descending price''}.}
\label{fig:demo_shopping}
\end{figure*}

\begin{figure*}[!ht]
\includegraphics[width=\textwidth]{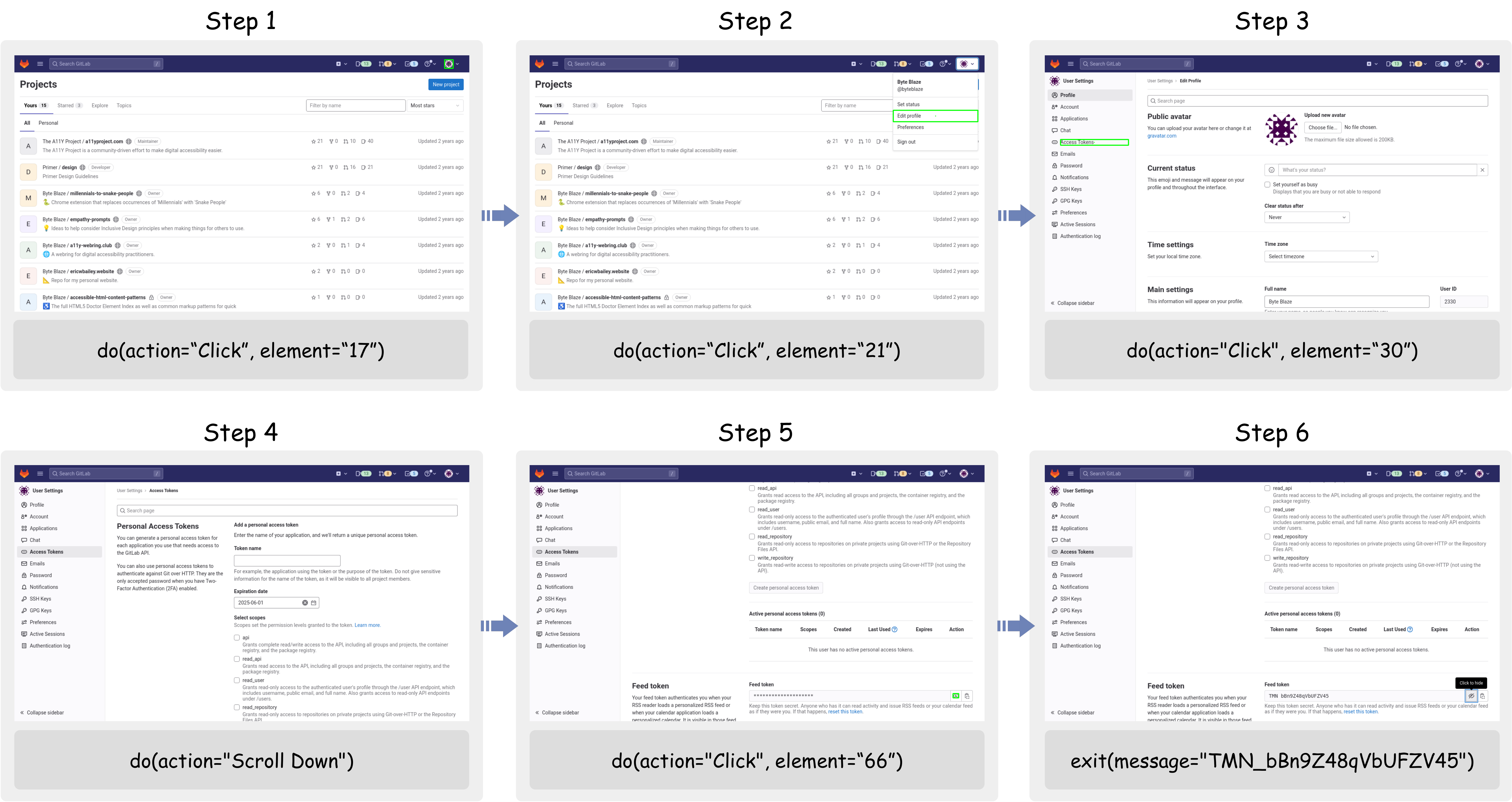}
\caption{A real-world example of a successful trajectory generated by \model on the GitLab task: \emph{``Get me my RSS feed token''}.}
\label{fig:demo_gitlab}
\end{figure*}

\begin{figure*}[!ht]
\includegraphics[width=\textwidth]{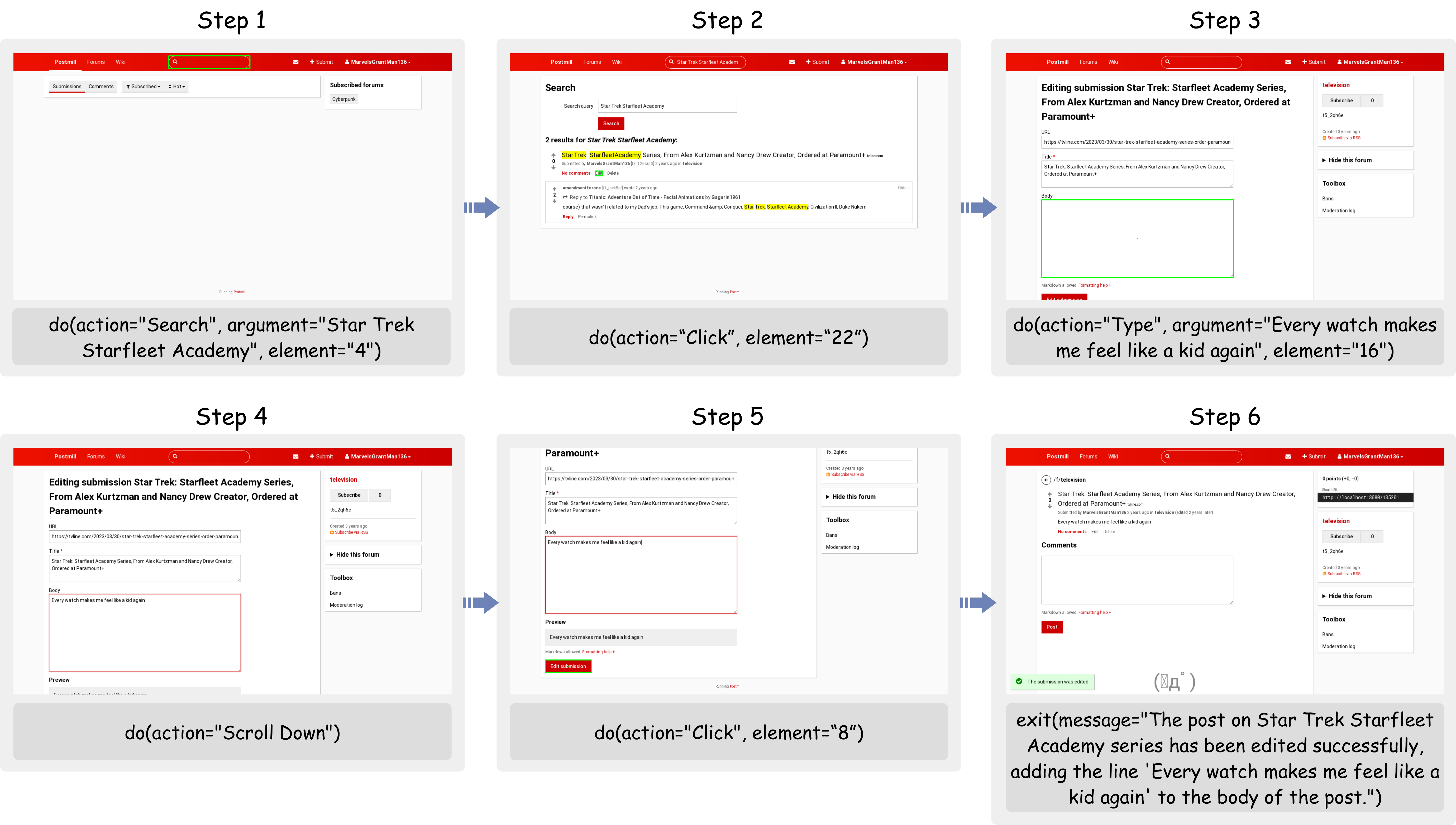}
\caption{A real-world example of a successful trajectory generated by \model on the Reddit task: \emph{``Edit my post on Star Trek Starfleet Academy series by adding a line to the body that says "Every watch makes me feel like a kid again"''}.}
\label{fig:demo_reddit}
\end{figure*}

In Figures~\ref{fig:demo_cms}–\ref{fig:demo_reddit}, we present a real-world successful trajectory generated by \model{} for each of the five websites in WebArena.

%CMS (Figure~\ref{fig:demo_cms}), Reddit (Figure~\ref{fig:demo_reddit}), GitLab (Figure~\ref{fig:demo_gitlab}), Map (Figure~\ref{fig:demo_map}), and Shopping (Figure~\ref{fig:demo_shopping}).

\begin{table*}[!ht]
\caption{Prompt used for long-CoT data augmentation, with example fill-ins for the fields \emph{user intent}, \emph{action history}, \emph{current observation}, \emph{next action}, and \emph{remarks}, all of which are available in the original behavior cloning data.
% Note that the \emph{observation} field in \emph{action history} is simplified using a fixed template to reduce context length as introduced in \S~\ref{sec:e2e_rl}.
The full HTML content in \emph{current observation} is omitted for clarity.}
\label{tab:data_aug_prompt}
\vspace{-1em}
\begin{prompt}[title={Long-CoT Data Augmentation Prompt}, label=prompt:data_aug_prompt]

You are an expert in explaining the behaviors of web agents.  
The agent is designed to help a human user navigate the website to complete a task.  
Given the user's intent, the agent's action history, the current HTML content of the web page, and the agent's next action associated with optional remarks,  
your goal is to explain the decision-making process from the agent's perspective using first-person narrative (as if the decision is being made in real time).\\

{\bf User Intent:}  
\texttt{``Establish a new discussion platform called `VirtualRealityVanguard'. It should serve as a cutting-edge forum where VR aficionados can engage in conversations about the newest trends, games, and applications in the virtual reality realm. Please ensure the sidebar features the following tags: virtual reality, technology, trends, gaming.''}\\

{\bf Action History}:  
\begin{verbatim}
 [
    {
      "round": "0",
      "observation": "** Simplified html **",
      "remarks": "# Element: the 'Forums' link at the top center",
      "action": "do(action="Click", element="1")"
    },
    {
      "round": "1",
      "observation": "** Simplified html **",
      "remarks": "# Element: the 'Create forum' button next to the 'List of forums' title",
      "action": "do(action="Click", element="17")"
    },
    {
      "round": "2",
      "observation": "** Simplified html **",
      "remarks": "# Element: the 'Name' field at the top of the page",
      "action": "do(action="Type", argument="VirtualRealityVanguard", element="12")"
    }
 ]
\end{verbatim}

{\bf Current Observation:} \texttt{<html> ... </html>}

{\bf Next Action:}
\texttt{do(action=``Type'', argument=``VirtualRealityVanguard'', element=``14'')}

{\bf Remarks:} \texttt{\# Element: the `Title' input field in the middle of the page}\\

Now, please explain the agent's thinking process using a first-person narrative.

\end{prompt}
\end{table*}

\begin{table*}[!ht]
\caption{System prompt for web agents. By default, we use the version with the thinking format (highlighted in gray). For the variant without the thinking format (discussed in \S~\ref{sec:analysis}), the gray part is simply removed.}
\label{tab:sys_prompt}
\vspace{-1em}
\begin{prompt}[title={System Prompt}, label=prompt:sys_prompt]
You are a professional web browsing agent assistant that can fulfill user's high-level instructions. Given simplified html of the browsed webpage at each step, you plan operations in python-style pseudo code using provided functions.\\

{\color{gray}You should first think about the reasoning process as an internal monologue and then decide an action. The reasoning process and answer are enclosed within \texttt{<think> </think>} and \texttt{<answer> </answer>} tags, respectively, i.e., responding in the following format: \texttt{<think> ... </think> <answer> ... </answer>}.}\\

More details about the code action: Your action should be readable, simple. Please generate **ONLY ONE ACTION** in one round. Predefined functions are as follows:

\begin{verbatim}
def do(action, argument, element):
    """A single browsing operation on the webpage.
    Args:
        :param action: one of the actions from ["Click", "Right Click", "Type", "Search", "Hover", 
                               "Scroll Up", "Scroll Down", "Press Enter", "Switch Tab", 
                               "Select Dropdown Option", "Wait"].
        :param argument: optional. Only for "Type", "Search", "Switch Tab", and
           "Select Dropdown Option", indicating the content to type in, page number (start from 0)
            to switch, or key to press. "Search" action is equivalent to "Type" action plus "Enter".      
        :param element: optional.  Only for "Click", "Right Click", "Type", "Search",
           "Select Dropdown Option", and "Hover". Should be specific element id in the HTML.
    Returns:
        None. The webpage will be updated after executing the action.
    """
    
def exit(message):
    """Ending the browsing process if the assistant think it has fulfilled the goal.
    Args:
        :param message: optional. If user's instruction is a question, return assistant's answer
           in the message based on the browsing content.
    Returns:
        None.
    """
    
def go_backward():
    """Go back to the previous page."""
    
def go_forward():
    """Go forward to the next page."""
\end{verbatim}

\textbf{Examples:}
\begin{itemize}[itemsep=2pt, parsep=0pt]
  \vspace{-0.5em}

\item \texttt{{\color{gray}<think>} \# Element: the 'REPORTS' section on the left sidebar {\color{gray}</think>}}\\
      \texttt{{\color{gray}<answer>} do(action="Click", element="7") {\color{gray}</answer>}}

\item \texttt{{\color{gray}<think>} \# Element: the 'Period' dropdown, middle center {\color{gray}</think>}}\\
      \texttt{{\color{gray}<answer>} do(action="Select Dropdown Option", argument="Month", element="20") {\color{gray}</answer>}}

\item \texttt{{\color{gray}<think>} \# Element: the 'From' date picker input field, middle center {\color{gray}</think>}}\\
      \texttt{{\color{gray}<answer>} do(action="Type", argument="01/01/2023", element="22") {\color{gray}</answer>}}

% \item \texttt{<think> reasoning process here </think>}\\
%       \texttt{<answer> do(action="Scroll Down") </answer>}

% \item \texttt{<think> reasoning process here </think>}\\
%       \texttt{<answer> exit(message="The top-3 best-selling products in January 2023 are: ...") </answer>}

% \item \texttt{<think> \# Element: The search bar </think>}\\
%       \texttt{<answer> do(action="Search", argument="international airport near Carnegie Mellon University within a driving distance of 50 km", element="13") </answer>}
\end{itemize}

\textbf{REMEMBER:}
\begin{itemize}[itemsep=2pt, parsep=0pt]
  \vspace{-0.5em}
  \item You can generate **ONLY ONE ACTION** in one round.
  \item If you have multiple potential actions to explore, you should generate other actions in separate rounds.
  \item Don't generate an operation element that you do not see in the screenshot.
  \item Use \texttt{``\# Element''} to describe the element you choose in the HTML.
  \item Use \texttt{``\# Note''} to record information useful to answer the instruction if needed.
  \item If you find yourself fallen into some sort of loop, try to use another method or change your action.
  \item If you think a page is still loading or still playing animation and you want to wait a while, use \texttt{``Wait''} action
  \item You are acting in a real world, try your best not to reject user's demand. Solve all the problem you encounter.
  \item If you think you didn't get expected webpage, you should try using more precise and locative description of the element.
  \item You should **NEVER** try to use the browser's address bar at the top of the page to navigate.
  \item Your answer shouldn't be in a code snippet format. Just write the function name and its arguments.
  \item If you use do function to perform \texttt{``Click''}, \texttt{``Right Click''}, \texttt{``Type''}, \texttt{``Search''}, \texttt{``Select Dropdown Option''}, and \texttt{``Hover''}, the parame \texttt{element} must not be \texttt{None}.
\end{itemize}
\end{prompt}
\end{table*}

\end{document}